\documentclass[letterpaper, 10 pt, journal, twoside]{ieeetran}  

\IEEEoverridecommandlockouts                              

\usepackage{graphicx} 
\usepackage{amsmath} 
\usepackage{amssymb}  
\usepackage{siunitx}
\usepackage{tikz}
\usepackage{tikz-3dplot}
\usepackage{lipsum}
\usepackage{cite}
\usepackage{hyperref}
\usepackage{multirow}
\usepackage[font=footnotesize,labelfont=bf]{caption}

\title{\LARGE \bf
Decentralized and Fully Onboard: Range-Aided Cooperative Localization and Navigation on Micro Aerial Vehicles
}

\author{Abhishek Goudar$^{1}$ and Angela P. Schoellig$^{1}$%
	\thanks{This work was supported in part by the Natural Sciences and Engineering Research Council of Canada (NSERC) and in part by the Canada CIFAR AI Chairs Program.}
	\thanks{$^{1}$The authors are with the Learning Systems and Robotics Lab
		at the Technical University of Munich, Germany, and the University of Toronto Institute for Aerospace Studies, Canada. They are also affiliated with the University of Toronto Robotics Institute, the Munich Institute of Robotics and Machine Intelligence (MIRMI), and the Vector Institute for Artificial Intelligence. E-mail: abhishek.goudar@robotics.utias.utoronto.ca, angela.schoellig@tum.de}
	\thanks{Digital Object Identifier (DOI): see top of this page.}
}

\DeclareMathOperator*{\argmax}{arg\,max}
\DeclareMathOperator*{\argmin}{arg\,min}

\begin{document}

\maketitle

\begin{abstract}
Controlling a team of robots in a coordinated manner is challenging because centralized approaches (where all computation is performed on a central machine) scale poorly, and globally referenced external localization systems may not always be available. In this work, we consider the problem of range-aided decentralized localization and formation control. In such a setting, each robot estimates its relative pose by combining data only from onboard odometry sensors and distance measurements to other robots in the team. Additionally, each robot calculates the control inputs necessary to collaboratively navigate an environment to accomplish a specific task, for example, moving in a desired formation while monitoring an area. We present a block coordinate descent approach to localization that does not require strict coordination between the robots. We present a novel formulation for formation control as inference on factor graphs that takes into account the state estimation uncertainty and can be solved efficiently. Our approach to range-aided localization and formation-based navigation is completely decentralized, does not require specialized trajectories to maintain formation, and achieves decimeter-level positioning and formation control accuracy. We demonstrate our approach through multiple real experiments involving formation flights in diverse indoor and outdoor environments.
\end{abstract}


\section{Introduction}

\IEEEPARstart{A}{}team of robots, micro aerial vehicles (MAVs) in this case, that interact in a collaborative manner to fly in a desired formation are suitable in applications that require a wide area coverage such as monitoring and search and rescue~\cite{chungSurveyAerialSwarm2018}. To execute such a formation flight, the robot team must be able to \textit{(i)} localize, i.e., estimate the pose of the individual robots and, \textit{(ii)} navigate a planned path in a pre-defined formation. Additionally, it is desirable to perform estimation and navigation in a decentralized and distributed setting (where the computation is done onboard the individual robots) as it is robust to a single point failure and can benefit from distributed storage and computation.

A common approach to cooperative localization is to use an infrastructure of known landmarks, such as the global positioning system (GPS)~\cite{bahrConsistentCooperativeLocalization2009} in outdoor environments or a network of ultra-wideband (UWB) radios~\cite{hamerSelfCalibratingUltraWidebandNetwork2018} in indoor environments. Infrastructure-based positioning may
not be feasible in many cases, as it is generally expensive and
unavailable in many environments. An alternative approach is
to use a combination of visual and inertial sensors to localize
against visual landmarks~\cite{tronDistributedOptimizationFramework2016}. Such an approach requires persistent visual landmarks and can be computationally expensive for real-time applications.

In this work, we propose a decentralized localization approach that uses range, visual, and inertial sensors to perform collaborative localization without the need for pre-installed infrastructure of sensors or building a persistent map of the environment. The combination of range, visual, and inertial sensors for decentralized estimation has been proposed before; however, the previous approaches generally require either strict coordination when solving the associated optimization problem or exchanging high-rate sensor data between agents to maintain consistent estimates. In contrast, our proposed approach has no such requirements. 

\begin{figure}[t]
	\centering
	\vspace*{0.5em}
	\includegraphics[scale=0.08,trim=0cm 15cm 0cm 30cm,clip]{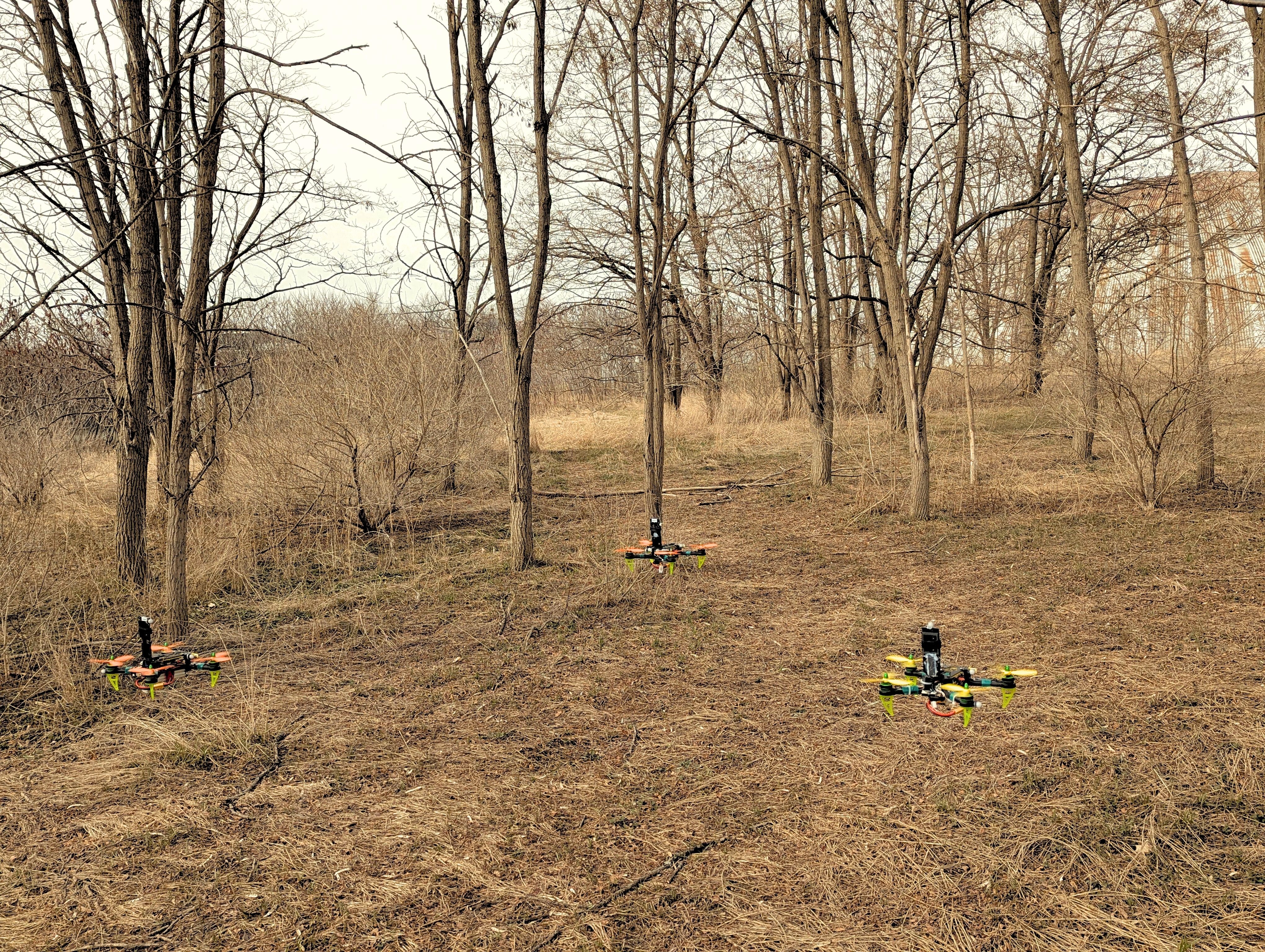}\\
	\vspace*{0.2em}
	\includegraphics[scale=0.32,trim=4cm 4cm 2.86cm 2cm,clip]{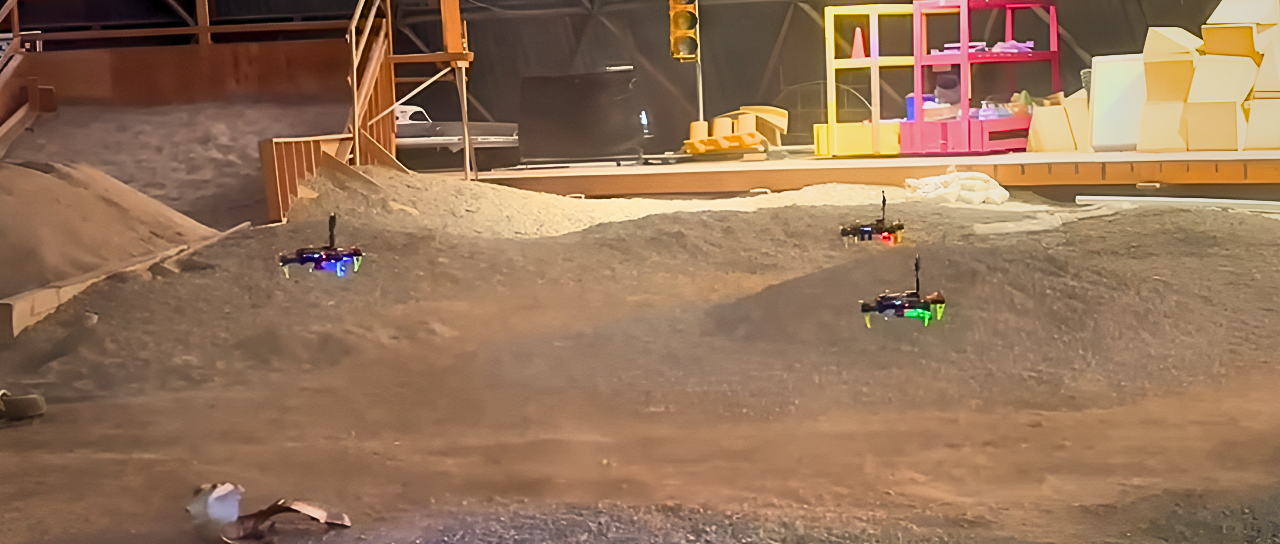}
	\caption{A snapshot of three micro aerial vehicles (MAVs) flying in a triangular formation under a tree canopy (top) and in a GPS-denied environment inside a metal dome (bottom) using our proposed method for localization and distance-based formation control. All estimation and control is done onboard the MAVs in a decentralized manner. A video of all the experiments can be found at following link: \textbf{\url{http://tiny.cc/mara_loc_nav}}. }
	\vspace*{-1.em}
	\label{fig:main_fig}
\end{figure}

The problem of coordinating multiple robots to move
in a desired formation is typically modeled as a leader-follower problem ~\cite{desaiControllingFormationsMultiple1998}, where the leader is commanded a pre-defined trajectory and the follower robots calculate control inputs to maintain the desired formation. The formation can be specified in terms of relative positions or distances. In this work, we consider distance-based formation control. Formation-based navigation has received considerable attention previously, however, in many cases, the effects of sensor noise and estimation uncertainty are not considered. For instance, off-the-shelf range sensors are noisy and affected by non-line-of-sight conditions, resulting in outliers and biased measurements. These non-idealities can result in poor tracking performance~\cite{goudarRangeVisualInertialSensorFusion2024}. In this work, we model formation planning as an inference problem on factor graphs~\cite{dellaertFactorGraphsExploiting2021} where we use continuous-time motion priors to generate smooth inputs for formation control. 
In summary, the contributions of our work are:
\begin{itemize}
	\item a novel approach to distributed distance-based formation control as an inference problem that is sparse, considers state uncertainty, and can be solved efficiently,
	\item a decentralized block coordinate descent approach to cooperative localization using range, visual, and inertial sensors that does not require exchanging odometry data or synchronous communication between agents, and 
	\item demonstration of joint localization and distance-based formation control on resource-constrained micro aerial vehicles in multiple diverse environments.
\end{itemize}

\section{Related Work}

In this section, we first review previous works that focus on decentralized cooperative localization, distance-based formation control, and joint localization and formation control. 

\subsubsection*{Decentralized localization} Our approach to decentralized cooperative localization is related to \textit{decentralized} state estimation. In~\cite{roumeliotisDistributedMultirobotLocalization2002}, distributed localization using a decomposed extended Kalman filter (EKF) with a densely connected network is shown. A distributed and decentralized approach to collaborative simultaneous localization and mapping (SLAM) using an EKF with distributed contributions is demonstrated in~\cite{leungDistributedDecentralizedCooperative2011}. While this approach recovers a globally-consistent estimate, it does so with delayed state estimates, which is not ideal for online control. Alternatively, a provably consistent approach to cooperative localization using covariance intersection (CI) in an EKF is demonstrated in~\cite{carrillo-arceDecentralizedMultirobotCooperative2013}. A combination of CI and odometry preintegration for collaborative localization, where each robot estimates the states of all its neighbors, is presented in ~\cite{cossetteDecentralizedStateEstimation2024}. In contrast, our proposed method does not communicate odometry information and each robot estimates its trajectory only.

Distributed estimation based on optimization-based approaches can benefit from improved accuracy by reducing linearization errors associated with filtering-based approaches. A distributed approach to localization based on maximum likelihood estimation (MLE), where each robot estimates its state assuming other robot states as constant, is shown in~\cite{howardLocalizationMobileRobot2003}. A maximum a posteriori (MAP) estimator that uses quantized measurements for efficient inference under communication constraints is demonstrated in~\cite{trawnyCooperativeMultirobotLocalization2009}. Similarly,~\cite{nerurkarDistributedMaximumPosteriori2009} proposes a MAP estimation approach to multi-robot cooperative localization that has convergence guarantees but requires synchronous communication, which can be difficult to obtain in real settings. A distributed and decentralized approach to SLAM is presented in~\cite{cunninghamDDFSAMConsistentDistributed2013a}, where consistency is enforced by subtracting double-counted information, but separate linearization between different robots is not accounted for. In this work, we use an asynchronous block coordinate descent (BCD) approach to MAP inference, that is decentralized and where each robot optimizes a local version of the MAP objective.

\subsubsection*{Formation control} Formation control is typically modeled using a graph theoretic-approach~\cite{yuRigidGraphControl2008}, where the formation is modeled as a graph and the desired control inputs are those that preserve the \textit{rigid} structure of the graph~\cite{krickStabilizationInfinitesimallyRigid2008}. Optimization-based decentralized control has been applied to high-speed formations flights with variable shapes ~\cite{turpinDecentralizedFormationControl2012}. 
%
An optimal control problem formulation for formation control is provided in~\cite{babazadehDistanceBasedFormationControl2021}, with extensions that incorporate unmodeled disturbances. In the previous cases, the effect of sensor noise and estimation uncertainty is not considered in formation control, which can be limiting when using range sensors with relatively large noise values. We present a probabilistic approach to formation control that takes the estimation uncertainty into account, and by extension the sensor noise. Factor graphs have been applied to multiple areas of robotics beyond state estimation, including planning~\cite{mukadamContinuoustimeGaussianProcess2018} and control~\cite{taFactorGraphApproach2014, dellaertFactorGraphsExploiting2021}. We extend the application of factor graphs to joint localization and formation control.

\subsubsection*{Cooperative localization and navigation} Next, we consider the body work on joint cooperative localization and navigation. The lightweight and inexpensive nature of visual and inertial sensors makes them ideal for multiagent formation control~\cite{tronDistributedOptimizationFramework2016, thakurSwarmInexpensiveHeterogeneous2021}. Reliance on maintaining visual line-of-sight to landmarks can be challenging and limiting in many cases. In~\cite{luskDistributedPipelineScalable2020}, formation control using onboard visual inertial odometry (VIO) and distributed formation execution is shown. Over time, VIO may accumulate drift and become inconsistent. A possible remedy is to generate a global consistent map~\cite{xuOmniSwarmDecentralizedOmnidirectional2022}, which can be computationally expensive and challenging in visually degraded conditions. Our proposed localization approach combines VIO with range sensors to obtain drift-free \textit{relative} state estimates that is computationally tractable and robust to visual occlusions. In~\cite{vanderhelmOnboardRangebasedRelative2020a}, a combination of range and inertial sensors is used for relative localization and formation control. A limitation of this estimation approach is that formation trajectories involving constant velocities are unobservable. Closely related to our work,~\cite{nguyenDistanceBasedCooperativeRelative2019, nguyenPersistentlyExcitedAdaptive2020,caoSimilarFormationControl2023} demonstrate joint localization and formation control using range, visual, and inertial sensors. However, these approaches require specialized trajectories involving persistent excitation to achieve formation, which our proposed method does not. More recently, impressive results involving decentralized navigation were presented in~\cite{zhouSwarmMicroFlying2022}, where the authors show position-based formation flights with obstacle avoidance. In contrast, we present an inference-based approach to distance-based formation control that incorporates  state uncertainty. Unlike~\cite{zhouSwarmMicroFlying2022}, we do not consider obstacle avoidance in this work.
\section{Problem statement}

The objectives of our work are to \textit{(i)} estimate the poses of a team of MAVs in a decentralized manner, and \textit{(ii)} generate smooth control inputs that consider state uncertainty to track a pre-specified distance-based formation pattern. An important objective of our work is to perform joint cooperative localization and navigation onboard the individual robots without centralized computing and coordination. We make the following assumptions. Each robot is equipped with a range sensor (such as a UWB radio), a camera, an inertial measurement unit (IMU), and an onboard compute unit. We assume that the desired formation is \textit{infinitesimally rigid}~\cite{yuRigidGraphControl2008}, specified in terms of inter-agent distances. Note that we do not assume a fully-connected graph for both estimation and formation control. We assume that each robot can communicate only with its immediate neighbors (and not the entire team) over a wireless network.
\section{Modeling}


\subsection{Decentralized Cooperative Localization}\label{sec:decentralized_cooperative_localization}

In this section, we provide a description of our proposed cooperative localization approach.
To represent information exchange, we use an undirected graph $\mathcal{G} = (\mathcal{V}, \mathcal{E})$, where $\mathcal{V} = \{ i\,| \, i \in \mathcal{S} \}$ is the set of vertices representing the robots as nodes in the graph, indexed using the set $\mathcal{S} = \{a, b, c, \hdots \}$, and $\mathcal{E} = \{(i,j) \,|\, i,j \in \mathcal{V}\}$ is the set of edges. An edge between a pair of vertices represents a range measurement between the corresponding agents.

We begin by describing the joint probability distribution over all the robot states and their measurements over a finite time window. Consider a team with $N = |\mathcal{S}|$ robots, where $|S|$ denotes the cardinality of set $S$. The state of a robot $a$ at time $t$ is given by its pose: $\mathbf{x}_{a_t} \in SE(d)$, where $d \in \{2,3\}$, $SE(d) \doteq \{(\mathbf{R}, \mathbf{p}) \, |\, \mathbf{p} \in \mathbb{R}^d, ~ \mathbf{R} \in SO(d) \}$ is the special Euclidean group, $\mathbf{p} \in \mathbb{R}^d$ is the position, and $\mathbf{R} \in SO(d)$ represents the rotation as member of the special orthogonal group. In this case, $\mathbf{x}_{a_t}$ refers to a $4\times4$ homogeneous matrix. The state of the team at some time $t$ is given by $\mathbf{x}_t = [\mathbf{x}_{a_t}^T, \mathbf{x}_{b_t}^T, \hdots, \mathbf{x}_{N_t}^T]^T$. We assume that each agent has two sensing modalities: \textit{(i)} an estimate of its ego motion obtained from VIO and \textit{(ii)} distance or range measurements to its neighboring agents.

The VIO algorithm estimates the pose of an agent relative to an arbitrary origin. Typically, these pose estimates are generated at rate higher than range measurements. One option is to include every pose estimate as a node in graph, however, this would result in increased computation cost with the size of the graph. We use \textit{odometry preintegration}~\cite{goudarRangeVisualInertialSensorFusion2024} where multiple VIO pose estimates are combined into a single relative pose between two time steps. We represent the VIO pose and the associated covariance matrix for an agent $a$ at time $t$ by the pair $(\bar{\mathbf{x}}_{o_{a_{t}}}, \boldsymbol{\Sigma}_{o_{a_{t}}})$. Consider a sequence of VIO pose estimates: $[(\bar{\mathbf{x}}_{o_{a_{t}}}, \boldsymbol{\Sigma}_{o_{a_{t}}}), (\bar{\mathbf{x}}_{o_{a_{t+1}}}, \boldsymbol{\Sigma}_{o_{a_{t+1}}}), \hdots, (\bar{\mathbf{x}}_{o_{a_{t'}}}, \boldsymbol{\Sigma}_{o_{a_{t'}}})]$. The consecutive pose estimates can be combined into a single \textit{relative} pose estimate $(\delta \bar{\mathbf{x}}_{o_{a_{t}}}, \boldsymbol{\Sigma}_{o_{a_{t}}}'')$ using the procedure outlined in~\cite{goudarRangeVisualInertialSensorFusion2024}. The single relative pose estimate relates the state between two distinct time instants $t$ and $t'$. It can be incorporated into the factor graph with the following model
\begin{equation}
	\delta\bar{\mathbf{x}}_{o_{a_{t}}} = (\mathbf{x}_{a_{t}}^{-1} \mathbf{x}_{a_{t'}}) \exp(\boldsymbol{\eta}_{{o}_{a_t}}^{\wedge}),  
\end{equation}
where $(\cdot)^\wedge$ maps a vector to the corresponding skew-symmetric matrix representing an element of the Lie algebra $\mathfrak{se}(d)$, $\exp(\cdot)$ is the matrix exponential that maps an element of $\mathfrak{se}(d)$ to $SE(d)$, and $\boldsymbol{\eta}_{{o}_{a_t}} \sim \mathcal{N}(\mathbf{0}, \boldsymbol{\Sigma}_{o_{a_{t}}}'')$. 

The generative model for the range measurement model is as follows. The distance measured by agent $a$ to agent $b$ at time $t$ is modeled as
\begin{equation}
	r_{{ab}_t} =  \| \bar{\mathbf{p}}_{{b}_t} - \bar{\mathbf{p}}_{{a}_t}  \|_2 + \eta_{{ra}_t},
	\label{eqn:inter_agent_range_measurement_model}
\end{equation}
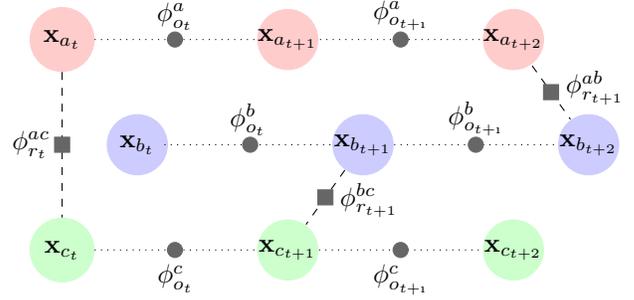
\begin{figure}[t]
	\hspace*{0.5em}
	\begin{tikzpicture}[scale=1.0]
	\node[circle, draw, fill,red!20,minimum size=0.85cm] (a0) at (0, 1.4) {};
	\node[] at (0, 1.4) {$\mathbf{x}_{a_{t}}$};
	\draw[fill,blue!20] (0+1, 0) circle (0.4cm);
	\node[] (b0) at (0+1, 0) {$\mathbf{x}_{b_{t}}$};
	\node[circle, draw, fill,green!20,minimum size=0.85cm] (c0) at (0, -1.4) {};
	\node[] at (0, -1.4) {$\mathbf{x}_{c_{t}}$};
	\foreach \i in {1,2} 
	{
		\draw[fill,red!20] (\i*3, 1.4) circle (0.4cm);
		\node[circle, inner sep=0, minimum size=0.1cm] (a\i) at (\i*3, 1.4) {$\mathbf{x}_{a_{t+\i}}$};
		\draw[fill,blue!20] (\i*3+1, 0) circle (0.4cm);
		\node[circle, inner sep=0, minimum size=0.1cm] (b\i) at (\i*3+1, 0) {$\mathbf{x}_{b_{t+\i}}$};
		\draw[fill,green!20] (\i*3, -1.4) circle (0.4cm);
		\node[circle, inner sep=0, minimum size=0.1cm] (c\i) at (\i*3, -1.4) {$\mathbf{x}_{c_{t+\i}}$};
	}
	{
		\draw[dotted] (a0) -- node[above]{ $\phi_{o_{t}}^a$} (a1) ;
		\draw[dotted] (b0) -- node[above]{ $\phi_{{o_{t}}}^b$} (b1);
		\draw[dotted] (c0) -- node[below, yshift=-2pt]{ $\phi_{{o_{t}}}^c$}(c1);
	}
	%
	{
		\draw[dotted] (a1) -- node[above]{ $\phi_{o_{t+\i}}^a$} (a2);
		\draw[dotted] (b1) -- node[above]{ $\phi_{{o_{t+\i}}}^b$} (b2);
		\draw[dotted] (c1) -- node[below, yshift=-2pt]{ $\phi_{{o_{t+\i}}}^c$}(c2);
	}
	\foreach \i in {1.5,4.5}  
	{
		\draw[fill,black!60] (\i, 1.4) circle (0.1cm);
		\draw[fill,black!60] (\i+1, 0) circle (0.1cm);
		\draw[fill,black!60] (\i, -1.4) circle (0.1cm);
	}
	{
		%
		\draw[dashed] (a0) -- node[anchor=east,xshift=-1pt]{$\phi_{{r_t}}^{ac}$} (c0);
		\draw[fill, black!60] (-0.1,-0.1) rectangle ++(0.2,0.2);
		
		\draw[dashed] (a2) -- node[anchor=south west, xshift=2pt,yshift=-8pt]{$\phi_{{r_{t+1}}}^{ab}$} (b2);
		\draw[fill, black!60] (6.4,0.6) rectangle ++(0.2,0.2);
		
		\draw[dashed] (b1) -- node[anchor=west,xshift=2pt,yshift=-1pt]{$\phi_{{r_{t+1}}}^{bc}$} (c1);
		\draw[fill, black!60] (3.4,-0.8) rectangle ++(0.2,0.2);
		
	}	
\end{tikzpicture}
	\caption{Factor graph for range-aided cooperative localization with three agents. The states of the agents at different time steps are shown using colored nodes. The edges between the nodes represent intra-agent and inter-agent measurements. Each agent measures its ego motion between using an odometry sensor represented by factor $\phi_{o_{t+\i}}^a = \phi_o(\mathbf{x}_{a_{t}}, \mathbf{x}_{a_{t+1}})$ (see~\eqref{eqn:odometry_factor_potential}). Additionally, an agent $a$ measures its distance to agent $b$ represented by factor $\phi_{{r_t}}^{ab} = \phi_r(\mathbf{x}_{a_{t}}, \mathbf{x}_{b_{t}})$.}
	\label{fig:multi_agent_est_factor_graph}
	\vspace*{-1.5em}
\end{figure}

where for an agent $a$, $\bar{\mathbf{p}}_{a_t} = \mathbf{R}_{a_t} \mathbf{p}_{l} + \mathbf{p}_{a_t}$ with $\mathbf{p}_{l}$ denoting the position offset between the IMU and the range sensor, and $\eta_{{ra}_t} \sim \mathcal{N}(0, \sigma_{r}^2)$ is the additive Gaussian noise. The factor graph for this setup is shown in Figure~\ref{fig:multi_agent_est_factor_graph}. The colored nodes in the graph represent robot states over time, the square factors between the nodes represent inter-agent range measurements, and the circular factors represent the preintegrated odometry factors. Assuming the Markov property over the states and independence of measurement noise, the joint probability distribution over the robots' states and measurements up to time step $K$ is 
\begin{align*}
	p(\mathbf{x}, \mathbf{y}) &=  \prod_{i \in \mathcal{S}} p(\mathbf{x}_{i_0} | \check{\mathbf{x}}_{i_0}) \prod_{i \in \mathcal{S}} \prod_{t=1}^{K-1} p(\mathbf{x}_{i_{t+1}} | \mathbf{x}_{i_{t}}, \delta \bar{\mathbf{x}}_{o_{{i}_{t}}})  \\ 
	& \prod_{i \in \mathcal{S}} \prod_{\substack{j \in \mathcal{S} \\ j \neq i}} \prod_{t=1}^{K} \kappa_{{ij}_t} ~ p(\mathbf{x}_{i_t}, \mathbf{x}_{j_t} | r_{{ij}_t})
\end{align*}
where $\mathbf{x} = [\mathbf{x}_0^T, \mathbf{x}_1^T,\hdots,\mathbf{x}_K^T]^T$, $\check{\mathbf{x}}_{i_0}$ is the known initial state, $\mathbf{y} = [\hdots,\delta \bar{\mathbf{x}}_{o_{{a}_{t}}}^T, \delta \bar{\mathbf{x}}_{o_{{b}_{t}}}^T,\hdots, r_{{ab}_t}, r_{{bc}_t},\hdots]^T$ is the set of all measurements, and $\kappa_{{ij}_t}$ is an indicator variable that is 1 if a range measurement exists between agents $a$ and $b$ at time $t$ and 0 otherwise. The robot state can be estimated by using MAP inference as
\begin{equation*}
	\mathbf{x}^* = \argmax_{\{\mathbf{x}_{i_t} \in SE(d)\}} p(\mathbf{x} | \mathbf{y}) \propto  \argmax_{\{\mathbf{x}_{i_t} \in SE(d)\}} p(\mathbf{x}, \mathbf{y}).
\end{equation*}

With the monotonicity of the (negative) logarithm function, the MAP inference objective simplifies to the following nonlinear least squares problem
\begin{align}
	J &= \min_{\{\mathbf{x}_{i_t} \in SE(d)\}} \Bigg( \sum_{i \in \mathcal{S}}  \phi_{\mathrm{init}}(\mathbf{x}_{i_0}) + \sum_{i \in \mathcal{S}} \sum_{t  = 1}^{K-1}  \phi_o(\mathbf{x}_{i_{t}}, \mathbf{x}_{i_{t+1}}) + \nonumber  \\
	& \sum_{i \in \mathcal{S}} \sum_{j \in \mathcal{S}} \sum_{t  = 1}^{K}
	\kappa_{{ij}_t} ~ \phi_r(\mathbf{x}_{i_t}, \mathbf{x}_{j_t})
	 \Bigg),
	 \label{eqn:swarm_map_obj}
\end{align}
where
\begin{align}
	\phi_{\mathrm{init}}(\mathbf{x}_{i_0}) &= \| \log(\check{\mathbf{x}}_{i_0}^{-1} \mathbf{x}_{i_0} )^{\vee} \|_{\check{\boldsymbol{\Sigma}}_{\check{\mathbf{x}}_{i_0}}}^2, \label{eqn:prior_factor_potential} \\ 
	\phi_o(\mathbf{x}_{i_{t}}, \mathbf{x}_{i_{t+1}}) &= \| \log( (\mathbf{x}_{i_t}^{-1} \mathbf{x}_{i_{t+1}})^{-1} \delta \bar{\mathbf{x}}_{o_{{i}_{t}}})^{\vee} \|_{{\boldsymbol{\Sigma}}_{o_{i_{t}}}''}^2, \label{eqn:odometry_factor_potential}\\
	\phi_r(\mathbf{x}_{i_{t}}, \mathbf{x}_{j_{t}}) &= \|r_{{ij}_t} -  (\| \bar{\mathbf{p}}_{{i}_t} - \bar{\mathbf{p}}_{{j}_t} \|_2) \|_{\sigma_{r}^2}^2, \label{eqn:range_factor_potential}
\end{align}
are the potential functions associated with different measurement factors, $\log (\cdot)$ is the matrix logarithm that maps an element of the Lie group to the corresponding Lie algebra, $(\cdot)^\vee$ maps a skew-symmetric matrix to a vector, $\check{\boldsymbol{\Sigma}}_{\check{\mathbf{x}}_{i_0}}$ is the uncertainty associated with initial state $\bar{\mathbf{x}}_{i_0}$,  and  $\| \mathbf{e}  \|_{\mathbf{Q}}^2 = \mathbf{e}^T \mathbf{Q}^{-1} \mathbf{e}$. The only term in the objective that depends on multiple agents is the range measurement factor~\eqref{eqn:range_factor_potential}; the initial state and odometry-based factors depend only on a single agent. Specifically, the joint optimization problem~\eqref{eqn:swarm_map_obj} can be decomposed into smaller sub-problems, where each agent only solves the part of the factor graph involving its states using the neighbor agent's pose and covariance. 


To solve~\eqref{eqn:swarm_map_obj} in a distributed manner, we adopt an asynchronous block coordinate descent (BCD)-based approach where each agent updates its state iteratively using the latest state estimates from its neighboring agents:
\begin{align}
	{\mathbf{x}_a^{k}} &= \argmin_{\{\mathbf{x}_{a_t} \in SE(d)\}} \left( \phi_{\mathrm{init}}(\mathbf{x}^{}_{a_0}) + \sum_{t  = 1}^{K-1}  \phi_o(\mathbf{x}_{a_{t}}, \mathbf{x}_{a_{t+1}}) + \nonumber \right. \\
	& \left. \sum_{j \in \mathcal{N}_{a_t}} \sum_{t  = 1}^{K-1}
	\kappa_{{aj}_t} ~ \tilde{\phi}_r(\mathbf{x}_{a_t}, \mathbf{x}_{j_t}^{k-1})  
	\right),
	\label{eqn:agent_bcd_iter}
\end{align}
where $\mathcal{N}_{a_{t}}$ is the set of neighbors of agent $a$ at time $t$ and $\mathbf{x}_{j_t}^{k-1}$ is state of agent $j$ from iteration $k-1$. The range factor $\tilde{\phi}_r$ is modified to account for the uncertainty in state $\mathbf{x}_{j_t}^{k-1}$: $\tilde{\phi}_r(\mathbf{x}_{i_{t}}, {\mathbf{x}}_{j_{t}}^{k-1}) = \|r_{{ij}_t} -  (\| \bar{\mathbf{p}}_{{i}_t} - \bar{\mathbf{p}}_{{j}_t}^{k-1} \|_2) \|_{\tilde{\sigma}_r^2}^2$, where $\tilde{\sigma}_r = \sigma_r + \mathrm{tr}(\boldsymbol{\Sigma}_{\check{\mathbf{x}}_{j_t}^{k-1}})$, where $\mathrm{tr}()$ is the matrix trace.

The number of neighbors, $ | \mathcal{N}_{a_t} |$, for any agent is dictated by the formation specification, which is presented in the next section. The nonlinear least squares problem associated with~\eqref{eqn:agent_bcd_iter} is solved using the Levenberg-Marquardt algorithm. The estimated state $\mathbf{x}_a^{k}$ is then shared with the neighboring agents to update their respective states $\{ \mathbf{x}_{j_t}^{k} \, | \, j \in \mathcal{N}_{a_t} \}$. To keep computational cost low, we use a \textit{fixed-lag smoother} approach, where states and measurements older than a particular fixed time window are marginalized. The iterations~\eqref{eqn:agent_bcd_iter} are run periodically at a fixed rate asynchronously; i.e., each agent solves an iteration of~\eqref{eqn:agent_bcd_iter} independently. The motivation for asynchronous BCD is to accommodate unreliable data networks, where solving~\eqref{eqn:swarm_map_obj} in coordinated manner cannot be guaranteed. We found in practice that asynchronous BCD achieves sufficient accuracy for reliable navigation. 


\subsection{Decentralized Formation Control}

The desired formation is represented using the graph $\mathcal{G}_d = (\mathcal{V}, \mathcal{E}_d)$ where $\mathcal{E}_d \subseteq \mathcal{E}$. The desired formation is parameterized by inter-agent distances and is assumed to be infinitesimally rigid. We formulate decentralized formation control as an instance of MAP inference that takes into account two sources of information: a motion prior that dictates the agent's motion over a finite time horizon and the distance constraints to neighboring agents for maintaining the desired formation.

We generate a motion prior for each agent using the framework of continuous-time Gaussian process (GP) regression~\cite{barfootStateEstimationRobotics2024}. The state $\boldsymbol{x}_{a_t} = [\boldsymbol{p}_{a_t}^T,\, \boldsymbol{u}_{a_t}^T] \in \mathbb{R}^{2d}$ consists of the agent's position $\boldsymbol{p}_{a_t} \in \mathbb{R}^d$ and the velocity control input $\boldsymbol{u}_{a_t} \in \mathbb{R}^d$. The motion model for an agent $a$ is:

\begin{align}
	\small
	\underbrace{
		\begin{bmatrix}
	 \dot{\boldsymbol{p}}_{{a}_t} \\
	 \dot{\boldsymbol{u}}_{{a}_t}
	\end{bmatrix}}_{\dot{\boldsymbol{x}}_{{a}_t}} &= 
	%
	%
	\underbrace{\begin{bmatrix}
		\mathbf{0} & \mathbf{I} \\
		\mathbf{0} & \mathbf{0}
	\end{bmatrix}}_{\mathbf{A}} 
	\underbrace{\begin{bmatrix}
		\boldsymbol{p}_{{a}_t} \\
		\boldsymbol{u}_{{a}_t}
	\end{bmatrix}}_{\boldsymbol{x}_{a_t}} + 
	\underbrace{\begin{bmatrix}
			\mathbf{0} \\
			\mathbf{I}
	\end{bmatrix}}_{\mathbf{L}}
	\boldsymbol{w}_{a_t},
	\label{eqn:wnoi_sys_model}
\end{align}
where $\mathbf{I}$ is the identity matrix and $\boldsymbol{w}_{a_t} \sim \mathcal{N}(0, \boldsymbol{Q}_{a_{\boldsymbol{w}}} \delta(t-t'))$ is a zero-mean GP with power spectral density matrix $\boldsymbol{Q}_{a_{\boldsymbol{w}}}$. The trajectory prior (mean and covariance) are calculated in closed-form~\cite[Section 3.4.2]{barfootStateEstimationRobotics2024} by integrating the motion model~\eqref{eqn:wnoi_sys_model} over a finite time horizon using the latest estimated state (and covariance) as the initial condition. The model~\eqref{eqn:wnoi_sys_model} is similar to a double-integrator model, but different in that it incorporates unmodeled disturbances with an additive noise term.  The mean of the GP prior between two consecutive time steps is
\begin{align}
	\boldsymbol{x}_{a_{t+1}} &= \boldsymbol{\Phi}_{t,t+1} \boldsymbol{x}_{a_{t}}
	\label{eqn:wnoi_mot_mean}
\end{align}
with
\begin{align*}
	\boldsymbol{\Phi}_{t,t+1} = \exp(\mathbf{A} \Delta t) &= \begin{bmatrix}
	\mathbf{I} & \Delta t \\
	\mathbf{0} & \mathbf{I}
	\end{bmatrix},
\end{align*}
where $\Delta t$ is the duration between time steps $t$ and $t+1$. The covariance associated with this incremental motion is
\begin{align}
	\mathbf{Q}_{a_t} &= \int_{t}^{t+1} \boldsymbol{\Phi}_{t,s} \mathbf{L} \boldsymbol{Q}_{a_{\boldsymbol{w}}} \mathbf{L}^T \boldsymbol{\Phi}_{s,t+1}^T ds, \\
	&= \begin{bmatrix}
		\frac{1}{3} \Delta t^{3} \boldsymbol{Q}_{a_{\boldsymbol{w}}} & \frac{1}{2} \Delta t^{2}\boldsymbol{Q}_{a_{\boldsymbol{w}}} \\
		\frac{1}{2} \Delta t^{2} \boldsymbol{Q}_{a_{\boldsymbol{w}}} & \Delta t \boldsymbol{Q}_{a_{\boldsymbol{w}}} 
	\end{bmatrix} \nonumber.
	\label{eqn:wnoi_mot_cov}
\end{align}

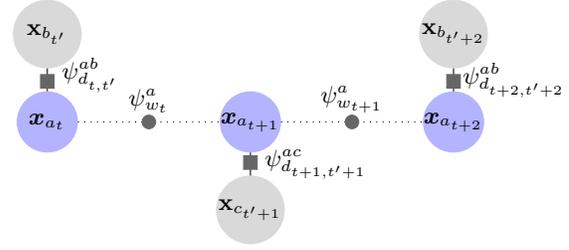
\begin{figure}
	\centering
	\begin{tikzpicture}[scale=0.9]
	\node[circle,draw,fill,gray!30, minimum size=0.9cm] (b1) at (0, 1.3){};
	\node[] at (0, 1.3) {\small$\mathbf{x}_{b_{t'}}$};
	\node[circle,draw,fill,gray!30, minimum size=0.9cm] (b2) at (6, 1.3){};
	\node[] at (2*3, 1.3) {\small$\mathbf{x}_{b_{t'+2}}$};
	\node[circle,draw,fill,blue!30, minimum size=0.8cm] (a0) at (0, 0){};
	\node[] at (0*3, 0) {\small$\boldsymbol{x}_{a_{t}}$};
	\foreach \i in {1,2} 
	{
		\node[circle,draw,fill,blue!30, minimum size=0.8cm] (a\i) at (\i*3, 0){};
		\node[]  at (\i*3, 0) {\small$\boldsymbol{x}_{a_{t+\i}}$};
	}
	\node[circle,draw,fill,gray!30, minimum size=0.9cm] (c1) at (3, -1.3){};
	\node[] at (3, -1.3) {\small$\mathbf{x}_{c_{t'+1}}$};
	%
	%
	\draw[dotted] (a0) -- node[above]{\small $\psi_{w_{t}}^a$} (a1);
	\draw[dotted] (a1) -- node[above]{\small $\psi_{w_{t+1}}^a$} (a2);
	\foreach \i in {1.5,4.5}  
	{
		\draw[fill,black!60] (\i, 0) circle (0.1cm);
	}
	{
		\draw[dashed] (a0) --  node[anchor=west,xshift=2pt, yshift=2pt]{\small$\psi_{d_{t,t'}}^{ab}$}(b1);
		\draw[fill, black!60] (-0.1,0.5) rectangle ++(0.2,0.2);
		\draw[dashed] (a1) -- node[anchor=west,xshift=2pt]{\small$\psi_{d_{t+1,t'+1}}^{ac}$} (c1);
		\draw[fill, black!60] (2.9,-0.7) rectangle ++(0.2,0.2);
		\draw[dashed] (a2) -- node[anchor=west]{\small$\psi_{d_{t+2,t'+2}}^{ab}$} (b2);
		\draw[fill, black!60] (5.9,0.5) rectangle ++(0.2,0.2);
		
	}
\end{tikzpicture}
	\caption{Factor graph associated with agent $a$'s formation control step. The state at a time $t$, $\boldsymbol{x}_{a_t}$, consists of the agent's position and velocity input. A Gaussian process-based smooth motion prior is added as a factor between two consecutive time stamps: $\psi_{w_{t}}^a = \psi_{w}(\boldsymbol{x}_{a_{t}}, \boldsymbol{x}_{a_{t+1}})$ (see~\eqref{eqn:form_ctrl_mm_factor_potential}). Additionally, the desired distances to neighbor agents, for maintaining a pre-specified formation, are represented by the factors: $\psi_{d_{t,t'}}^{ab} = \psi_d(\boldsymbol{x}_{a_{t}}, \mathbf{x}_{b_{t'}})$ (see~\eqref{eqn:form_ctrl_dist_factor_potential}), where the neighboring agent's state is fixed to its latest estimate (as indicated by the grayed out node).}
	\vspace*{-1.5em}
	\label{fig:formation_factor_graph}
\end{figure}
In addition to the motion model, at each time step, agent $a$ must also maintain specific distance to its neighbors in order to achieve the desired formation. One approach is to incorporate this distance requirement as \textit{constraints} in the optimization problem~\cite{zhangEfficientRangeConstraintManifold2022}. We consider an unconstrained formulation where the desired distances are incorporated into the cost function with	 the following generative model:
\begin{align}
	d_{aj_t} = \| \boldsymbol{p}_{a_t} - \mathbf{p}_{j_{t'}} \|_2 + \eta_{{j}_{t'}}, \quad j \in \mathcal{N}_{a_t},
\end{align}
where
$\mathbf{p}_{j_{t'}}$ is the estimated position of agent $j$ at some time $t'$ (typically older than $t$). The additive noise $ \eta_{{j}_{t'}} \sim \mathcal{N}(0, Q_{{j_{t'}}})$ takes into account the marginal uncertainty of agent $j$'s position: $Q_{{j_{t'}}} = \mathrm{tr}(\hat{\boldsymbol{\Sigma}}_{\mathbf{p}_{j_{t'}}})$, where $\hat{\boldsymbol{\Sigma}}_{\mathbf{p}_{j_{t'}}}$ is the sub-matrix corresponding to the estimated uncertainty in position. The motivation for our proposed formulation is as follows. Firstly, the motion prior of an agent is generated using its own estimation uncertainty and its motion model uncertainty. This has the effect of generating smooth priors when the estimation uncertainty or model uncertainty is high. Secondly, we use the estimated states of the neighboring agents $\mathbf{p}_{j_{t'}}$, instead of their predicted states, as using predicted states would require exchanging information at the formation controller frequency, which can be expensive and cannot be guaranteed in practice. Thirdly, our approach incorporates estimation uncertainty of neighboring agents as well: if an agent has high state uncertainty, the corresponding distance constraint gets penalized more in the optimization problem, which is desirable. Finally, we incorporate the desired inter-agent distances in the cost function (instead of as equality constraints) as it may not be possible to strictly adhere to inter-agent distances for short periods of time in lieu of unreliable communication network or other critical tasks.

The factor graph associated with the formation controller of a single agent is shown in Figure~\ref{fig:formation_factor_graph}. Note that the structure of the formation control problem is similar to that of the localization problem from the previous section. Agent $a$'s objective function for formation control is as follows:
\begin{align}
	\tilde{J}_{a} &= \min_{\{\boldsymbol{x}_{a_t} \in \mathbb{R}^{2d}\}} \left(  \psi_{\mathrm{init}}(\boldsymbol{x}_{a_K}) + \sum_{t  = K}^{K'-1}  \psi_{w}(\boldsymbol{x}_{a_{t}}, \boldsymbol{x}_{a_{t+1}}) + \nonumber \right. \\
	& \left. \sum_{j \in \mathcal{S}} \sum_{t  = 1}^{K'}
	\kappa_{{aj}_t} ~ \psi_d(\boldsymbol{x}_{a_t}, \mathbf{x}_{j_{t'}})  
	\right),
	\label{eqn:form_ctrl_obj}
\end{align}
where $K' > K$, $\boldsymbol{x}_a = [ \boldsymbol{x}_{a_K}^T, \boldsymbol{x}_{a_{K+1}}^T, \hdots, \boldsymbol{x}_{a_{K'}}^T]^T$,
\begin{align}
	\psi_{\mathrm{init}}(\boldsymbol{x}_{a_K}) &= \| \bar{\boldsymbol{{x}}}_{a_K} - \boldsymbol{x}_{a_K} \|_{\mathbf{Q}_{\mathbf{x}_{a_K}}}^2, \label{eqn:form_ctrl_init__potential} \\ 
	\psi_{w}(\boldsymbol{x}_{a_{t}}, \boldsymbol{x}_{a_{t+1}}) &= \| 
	\boldsymbol{x}_{a_{t+1}} - \boldsymbol{\Phi}_{t,t+1} \boldsymbol{x}_{a_t}
	 \|_{\mathbf{Q}_{a_t}}^2, \label{eqn:form_ctrl_mm_factor_potential}\\
	\psi_d(\boldsymbol{x}_{a_{t}}, \mathbf{x}_{j_{t'}}) &= \|d_{{aj}_t} -  (\| {\boldsymbol{p}}_{{a}_t} - {\mathbf{p}}_{j_{t'}} \|_2) \|_{Q_{{j_{t'}}}}^2, \label{eqn:form_ctrl_dist_factor_potential}
\end{align}
and $\bar{\boldsymbol{x}}_{{a}_K} = [\mathbf{p}_{a_K}^T, \mathbf{v}_{a_K}^T]^T$
consists of the estimated position $\mathbf{p}_{a_K}$, and the velocity estimate $\mathbf{v}_{a_K}$ from odometry. The uncertainty matrix $\mathbf{Q}_{\mathbf{x}_{a_K}}$ is a block diagonal matrix composed of the marginal covariance of agent $a$'s estimated position uncertainty $\hat{\boldsymbol{\Sigma}}_{\mathbf{p}_{a_t}}$ and the uncertainty associated with velocity reported by the VIO algorithm $\mathbf{Q}_{v_K}$: $\mathbf{Q}_{\mathbf{x}_{a_K}} = \mathrm{diag}(\hat{\boldsymbol{\Sigma}}_{\mathbf{p}_{a_t}}, \mathbf{Q}_{v_K})$, where the $\mathrm{diag}(\cdot)$ operator constructs a block-diagonal matrix from its arguments. As before, $\kappa_{{aj}_t}$ is an indicator variable that is 1 if a distance constraint exists between agents $a$ and $i$ at time $t$ and 0 otherwise. The above formulation takes the estimated state uncertainty into account in the factor $\psi(\boldsymbol{x}_{a_K})$. The desired positions and control inputs to achieve the desired formation are obtained by minimizing~\eqref{eqn:form_ctrl_obj}:
\begin{align}
	\boldsymbol{x}_a^{*} &= \argmin_{\{ \boldsymbol{x}_{a_t} \in \mathbb{R}^{2d} \}} \tilde{J}_a.
	\label{eqn:form_ctrl_agent_state}
\end{align}

Solving~\eqref{eqn:form_ctrl_agent_state} gives agent $a$'s positions and the corresponding velocity control inputs over a finite time horizon. We can view the calculated state $\boldsymbol{x}_{a_t}^* = [{\boldsymbol{p}_{a_t}^*}^T,\, {\boldsymbol{u}_{a_t}^*}^T]^T$ as consisting of a position control input and a velocity control input to achieve the desired formation. Depending on the lower-level control interface, either input can be chosen. We observed that the position control inputs resulted in smoother control as the velocity inputs were more susceptible to errors in presence of higher measurement noise. To keep the computational cost low, we solve the optimization problem~\eqref{eqn:form_ctrl_agent_state} in a fixed-lag smoother fashion over a finite time horizon. We use only the last control input $\boldsymbol{x}_{a_{K'}}^*$ and the optimization problem is run periodically at a fixed rate. Each agent independently executes its version of~\eqref{eqn:form_ctrl_agent_state} for formation control. 

\subsection{Implementation Details}
The asynchronous BCD approach does not require strict coordination while solving the back-end optimization problem. However, network delays and packet losses can cause measurements to be delayed or missed entirely. Note that the only data exchanged between the agents are the range measurements and the corresponding agent poses and covariances. To delay with network delays and packet losses, we use the dual smoother architecture outlined in~\cite{goudarRangeVisualInertialSensorFusion2024}. Specifically, the FLS associated with the decentralized estimator runs at a lower rate and does not require measurements to arrive in real time. Concretely, if a range measurement is delayed due to network latency, it is added to the factor graph at the corresponding time step in the past. Similarly, if a neighbor agent does not respond with its pose for a particular range measurement or the corresponding packet is lost, the range measurement is removed from the factor graph. To facilitate closed-loop navigation, the low-rate pose output of the decentralized estimator is then combined with the latest VIO estimates in another smoother that runs at a higher rate for closed-loop navigation. 

\section{Evaluation}

We evaluate our proposed cooperative localization and navigation approach through \textit{leader-follower} formation flights in simulation and real experiments. In this case, only the leader agent is commanded a pre-defined trajectory, the other agents (followers) compute control inputs to follow the leader while maintaining a desired formation. For evaluation, we assume that only the leader has a source of absolute pose to track its designated trajectory. Note that this assumption is for evaluation only and can be relaxed as shown in the accompanying video. The localization and formation control algorithms are implemented using the GTSAM library~\cite{gtsam}. Inter-agent communication is done over WiFi using ROS2 Jazzy (with eProsima Fast DDS backend) middleware. 

We compare our proposed localization approach against centralized batch trajectory estimation (referred to as \textit{batch}) as a baseline. To evaluate the localization performance, we use the average position \textit{root mean square error} (RMSE) of all the follower agents as a metric. 
We compare our proposed formation control approach with the gradient control method (referred to as \textit{gcm}) outlined in \cite{krickStabilizationInfinitesimallyRigid2008}, which is a decentralized approach to rigidity-based formation control. The formation control methods are evaluated by comparing the desired and observed distances using the \textit{formation root mean square error} (FRE) metric:
\begin{align}
	\mathrm{FRE} \doteq \frac{1}{K}  \sqrt{\sum_{t=1}^K ( \bar{d}_{ij_{t}} - d_{ij_{t}})^2},~(i,j) \in \mathcal{E}_d
	\label{eqn:formation_rmse_error}
\end{align}
where $\bar{d}_{ij_{t}}$ and $d_{ij_{t}}$ are the desired and observed distances between agents $i$ and $j$ at time $t$, respectively. 


\subsection{Simulation}

Our simulation setup is based on Ignition Gazebo and consists of a generic environment with multiple MAVs. Each MAV has \textit{(i)} a range sensor and a generic odometry sensor (to simulate VIO) for estimation, and \textit{(ii)} a low-level controller to convert high-level setpoints (position and velocity) into motor inputs for closed-loop control. The sensor data is corrupted by adding Gaussian noise reflective of real sensors: $\sigma_r = 0.05\,\si{m}$. The desired formation is specified a priori. The simulations are run on a single laptop with the autonomy stack (including estimation and formation control) associated with different MAVs running in a decentralized manner.

\begin{figure}[t]
	\hspace{-0.5em}
	\includegraphics[scale=0.78]{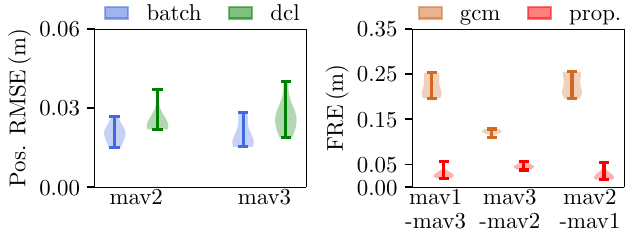}
	\vspace*{-0.5em}
	\caption{Error distribution plots for localization (left) and formation control (right) from 10 simulations. The proposed decentralized cooperative localization (dcl) method has positioning RMSE similar to centralized batch trajectory estimation. The formation RMSE (FRE) (see~\eqref{eqn:formation_rmse_error}) distributions show that the proposed (prop.) approach achieves lower formation error than the baseline gradient control (gcm) approach.}
	\label{fig:sim_rmse_results}
	\vspace*{-1em}
\end{figure}

\begin{figure}[t]
	\centering
	\includegraphics[scale=0.74,trim={0 0.3cm 0 0}]{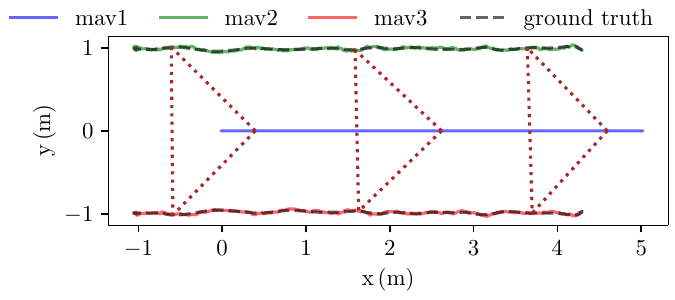}
	\caption{Estimated and ground-truth trajectories from a leader-follower flight in simulation. The leader (mav1) tracks a straight path, and the other agents (mav2 and mav3) estimate their state and the control inputs to follow the leader in a triangular formation (indicated by the red dotted lines).}
	\vspace*{-1.8em}
	\label{fig:sim_traj_results}
\end{figure}

\begin{figure}[t]
	\centering
	\includegraphics[scale=0.79]{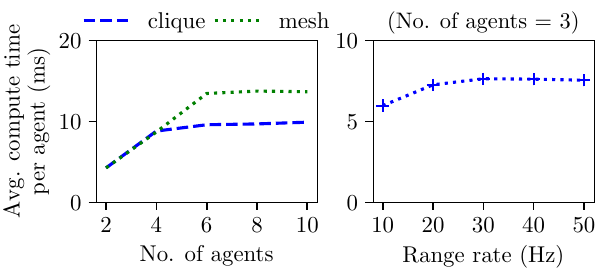}
	\vspace*{-0.3em}
	\caption{Average localization solve times per agent vs. (left) number of neighbor agents under mesh (dense) and clique (sparse) connectivity, and (right) range update rate with a team of three agents.}
	\label{fig:agt_cnt_avg_slv_time}
	\vspace*{-1.5em}
\end{figure}

One of the MAVs is designated as a leader and commanded predefined trajectories; the follower agents use the proposed method to estimate their state and generate control inputs to maintain the desired formation. To prevent the team collectively translating along $z$-axis (potentially flying into the ground plane), we add additional constraints to the formation control factor graph to maintain constant altitude relative to the terrain. Quantitative results showing estimation and formation RMSE with three MAVs flying in a triangular formation from 10 trials are shown in Figure~\ref{fig:sim_rmse_results}. Qualitative results from one such experiment showing the estimated trajectories are shown in Figure~\ref{fig:sim_traj_results}. The proposed fixed-lag smoother-based decentralized cooperative localization (dcl) has positioning RMSE similar to centralized batch trajectory estimation. As highlighted in~\cite{goudarRangeVisualInertialSensorFusion2024}, with noisy range measurements, the estimated pose tends to be non-smooth.  The estimation uncertainty captures this variation in the estimated pose, which is subsequently used in our formation control approach. As such, the setpoints from our formation control pipeline afforded smoother trajectory tracking. The baseline approach does not account for the estimation uncertainty and results in higher formation RMSE (see Figure~\ref{fig:sim_rmse_results}) and unreliable control. This aspect is also highlighted in the accompanying video. Note that the the formation error magnitude is influenced by the estimation error, since we are doing closed-loop navigation.

To demonstrate the scalability of our method, we computed the average time taken by a single agent to solve its localization and formation control sub-problems with increasing number of robots. Since our approach is decentralized, we focus on the computation time of a single agent. To evaluate real-world scalability, we ran the localization and the formation control pipelines associated with one of the simulated agents on the (computationally-constrained) single board computer used in our real experiments. The rest of the simulated agents (and their processes) were run on a laptop. The range update rate is fixed to $10\,\si{Hz}$. We evaluated two communication paradigms: \textit{(i)} clique, where each agent communicates with a maximum of four neighbors, and \textit{(ii)} mesh, where each agents communicates with every other in the team. We stress here that such a mesh network is not needed for our proposed method, but presented here as the worst-case communication scenario. The average compute time with increasing number of agents in shown in Figure~\ref{fig:agt_cnt_avg_slv_time} (left). The information exchange between the agents is governed by the range update rate. The average compute time of a single agent (in a group of three agents) with increasing range update rate is shown in Figure~\ref{fig:agt_cnt_avg_slv_time} (right). The results show that our approach is scalable as the compute time does not increase significantly with increasing number of agents. Compute times for formation control follow a similar trend. Qualitative results showing multiple formations with varying number of agents can be found in the accompanying video.

\subsection{Real-world Experiments}


\begin{figure}[t]
	\includegraphics[scale=0.78,trim={0 0 0 0}, clip]{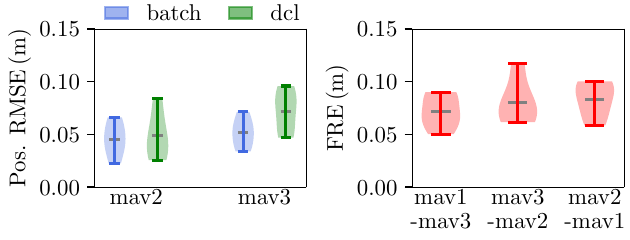}

	\vspace*{-0.3em}
	\caption{Localization (left) and formation (right) error distributions from five real experiments where the leader is commanded a straight line path. The proposed decentralized cooperative localization (dcl) method has positioning RMSE close to the centralized batch trajectory estimation. The proposed formation control approach achieves the desired formation as indicated by the formation RMSE (FRE) distributions.}
	\label{fig:real_rmse_results}
	\vspace*{-2em}
\end{figure}

The MAVs used in our experiments are equipped with a Qualcomm Flight Pro single board computer (Quad-core 2.15~G\si{Hz} CPU with 4\,GB RAM) with an onboard camera and an IMU, a Loco-Positioning UWB node ($\sigma_r = 0.1\si{m}$), and a Pixracer flight control unit (FCU) running the PX4-Autopilot firmware. We use UWB radios in two-way ranging (TWR) mode, where only two agents can measure distance to each other at any given time. To facilitate range measurements across multiple agents, we use time division to share the UWB channel: each agent operates its UWB radio in a pre-specified time slot to measure the distance to its neighbors. The range measurement rate for the entire team is set to 12\,\si{Hz}. The onboard VIO generates pose estimates at 30\,\si{Hz}. As before, for safety, we use a downward-facing altimeter to enforce constant altitude constraint to avoid collisions with uneven terrains and bushes in the environment. We use setpoints calculated using our proposed method, that are subsequently sent to the FCU's onboard controller.
Each MAV runs the necessary processes, including the localization and formation control pipelines, on the onboard computer. The sensor data is collected for offline comparison with the batch approach. We evaluated our approach through 22 flight experiments in three different environments. Our proposed method was able to successfully execute formation flights in all of the experiments. 

\begin{figure}[t]
	\hspace*{-0.8em}
	\includegraphics[scale=0.61, trim={0 0.2cm 0 0.2cm}]{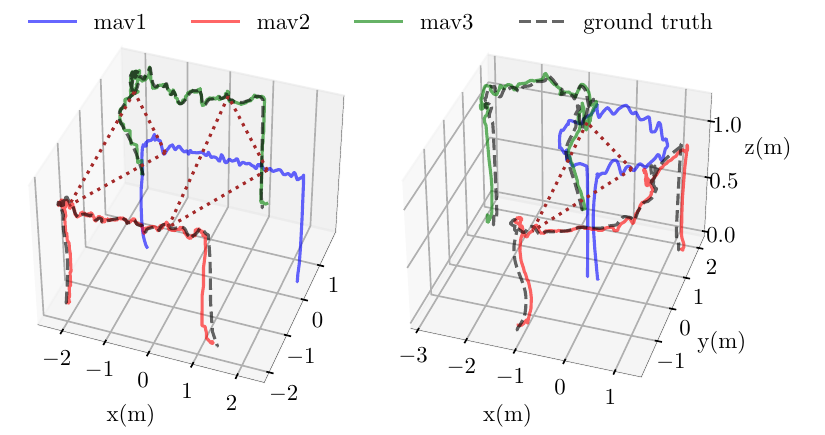}
	\caption{Estimated and ground-truth trajectories from real flights. The leader agent (mav1) tracks a straight path (left) and a circular path (right), and the other agents (mav2 and mav3) estimate their state and the control inputs to follow the leader in a triangular formation (shown by the red dotted lines).}
	\label{fig:real_traj_results}
	\vspace*{-1.6em}
\end{figure}

Our first test environment is a flight arena equipped with a motion capture system, which is used as the source of ground-truth position. We performed multiple leader-follower flights where the leader agent is commanded a straight path and a circular path. We discuss results for the straight path first. The leader agent (mav1) uses the motion capture system for tracking the trajectory its commanded, while the follower agents (mav2 and mav3) estimate their state using our proposed method. The follower agents calculate the necessary control inputs to maintain a pre-specified triangular formation. 
Box plots showing position RMSE from five such leader-follower flights are shown Figure~\ref{fig:real_rmse_results}. Qualitative results showing a subset of the estimated trajectories are shown in Figure~\ref{fig:real_traj_results} (left). We see that our proposed decentralized cooperative localization (dcl) approach achieves positioning RMSE similar to centralized (batch) trajectory estimation. Note that the proposed dcl method uses only a subset of measurements in the fixed-lag smoother, where as the batch approach uses the entire history of measurements. These results are promising as the UWB radios used in our experiments have a range precision of $\pm\,0.1\,\si{m}$. The bounded formation RMSE in Figure~\ref{fig:real_rmse_results} shows that the proposed method is able to maintain the desired formation. The maximum linear speed of the agents in these experiments was $0.4~\si{m/s}$. Note that our proposed formation control approach is reactive in nature: each agent calculates the necessary control inputs after receiving states from its neighbors. As such, computation and communication-related delays result in larger tracking error at higher speeds. Qualitative results demonstrating our approach at higher speeds can be found in the accompanying video. Unlike in simulation, we found that in real experiments, the baseline formation controller (gcm) was very susceptible to computation and communication-related delays, which resulted in unreliable tracking behavior.

\begin{table}[b]
	\vspace*{-1.5em}
	\caption{Localization and formation control RMSE from three experiments where the leader is commanded a circular trajectory. (Left) The mean and standard deviation (std) of positioning RMSE from the proposed approach (dcl) are comparable to the centralized (batch) approach. (Right) The formation RMSE (FRE) indicates that the proposed approach achieves the desired formation.}
	\resizebox{0.495\columnwidth}{!}
	{%
		\addtolength{\tabcolsep}{-0.6em}
		\begin{tabular}{ | c | c | c |} 
			\hline
			\multicolumn{3}{|c|}{Position RMSE}\\			
			\hline
			& \multicolumn{2}{c|}{mean $\pm$ std ($\si{m}$)}\\
			\cline{2-3}
			& dcl & batch \\
			\hline
			mav2 & 0.10 $\pm$ 0.01 & \textbf{0.09 $\pm$ 0.02}  \\ 
			\hline
			mav3 & 0.10 $\pm$ 0.01 & 0.10 $\pm$ 0.03\\
			\hline
		\end{tabular}
	}
	\resizebox{0.48\columnwidth}{!}
	{%
		\addtolength{\tabcolsep}{-0.5em}
		\begin{tabular}{ | c | c |} 
			\hline
			\multicolumn{2}{|c|}{Formation RMSE} \\
			\hline
			& {  mean $\pm$ std ($\si{m}$)}\\
			\cline{1-2}
			mav1-mav2 & 0.20 $\pm$ 0.02 \\
			\hline
			mav1-mav3 & 0.22 $\pm$ 0.03 \\ 
			\hline
			mav2-mav3 & 0.10 $\pm$ 0.01 \\
			\hline
		\end{tabular}
	}
	\label{tab:circ_path_rmse}
\end{table}

Next, we discuss results from the second set of experiments where the leader agent is commanded a circular path. The average localization and formation RMSE from three such experiments are shown in Table~\ref{tab:circ_path_rmse}. Qualitative results showing the estimated trajectories from one such experiment are shown in Figure~\ref{fig:real_traj_results} (right). The localization RMSE is similar to the previous case, however, the formation RMSE is on the higher end. Additionally, with a circular trajectory, minor oscillations in the leader's position can lead to bigger oscillations in other agents' positions for a larger radius.

Qualitative results for the experiments below can be found in the accompanying video. The high-bandwidth nature of UWB allows measuring distances across obstacles, such as foam, cardboard, and other non-metallic objects, where visual line of sight (LOS) is obstructed but radio LOS persists. We found that the foam and cardboard had minimal impact on the ranging accuracy. Our method can execute formation flights in such settings as well. Our second test environment is a metal dome, where three MAVs collaboratively navigate in a triangular formation to inspect a simulated crash site. This environment is challenging as there is no GPS (see Figure~\ref{fig:main_fig}). We also evaluated our approach in an outdoor environment under a tree canopy (where GPS can be unreliable) in the presence of wind disturbances. 

\vspace*{-0.5em}

\section{Conclusion}

In this work, we presented an approach to joint range-aided cooperative localization and distance-based formation control that is truly decentralized and distributed. We validated our approach through real experiments involving resource-constrained micro aerial vehicles executing formation flights in multiple diverse environments. The results show that our proposed localization and formation control methods achieve decimeter-level positioning and formation accuracy. For future work, we plan to investigate increasing the speed of the formation flights and incorporating obstacle avoidance in the factor graph framework.

\bibliographystyle{unsrt}
\bibliography{main_reference.bib}

@book{barfootStateEstimationRobotics2024,
  title = {State {{Estimation}} for {{Robotics}}: {{Second Edition}}},
  shorttitle = {State {{Estimation}} for {{Robotics}}},
  author = {Barfoot, Timothy D.},
  year = {2024},
  month = jan,
  publisher = {Cambridge University Press},
  doi = {10.1017/9781009299909},
}

@article{chungSurveyAerialSwarm2018,
	title = {A {{Survey}} on {{Aerial Swarm Robotics}}},
	author = {Chung, Soon-Jo and Paranjape, Aditya Avinash and Dames, Philip and Shen, Shaojie and Kumar, Vijay},
	year = {2018},
	month = aug,
	journal = {IEEE Transactions on Robotics},
	volume = {34},
	number = {4},
	pages = {837--855},
	issn = {1552-3098, 1941-0468},
	doi = {10.1109/TRO.2018.2857475},
}

@article{hamerSelfCalibratingUltraWidebandNetwork2018,
	title = {Self-{{Calibrating Ultra-Wideband Network Supporting Multi-Robot Localization}}},
	author = {Hamer, Michael and Dandrea, Raffaello},
	year = {2018},
	journal = {IEEE Access},
	volume = {6},
	pages = {22292--22304},
	publisher = {IEEE},
	issn = {21693536},
	doi = {10.1109/ACCESS.2018.2829020}
}

@inproceedings{bahrConsistentCooperativeLocalization2009,
	title = {Consistent Cooperative Localization},
	booktitle = {Proc. of the {{International Conference}} on {{Robotics}} and {{Automation}} ({ICRA})},
	author = {Bahr, A. and Walter, M.R. and Leonard, J.J.},
	year = {2009},
	month = may,
	pages = {3415--3422},
	publisher = {IEEE},
	address = {Kobe},
	doi = {10.1109/ROBOT.2009.5152859},
	isbn = {978-1-4244-2788-8},
}

@article{tronDistributedOptimizationFramework2016,
	title = {A {{Distributed Optimization Framework}} for {{Localization}} and {{Formation Control}}: {{Applications}} to {{Vision-Based Measurements}}},
	shorttitle = {A {{Distributed Optimization Framework}} for {{Localization}} and {{Formation Control}}},
	author = {Tron, Roberto and Thomas, Justin and Loianno, Giuseppe and Daniilidis, Kostas and Kumar, Vijay},
	year = {2016},
	month = aug,
	journal = {IEEE Control Systems},
	volume = {36},
	number = {4},
	pages = {22--44},
	issn = {1066-033X, 1941-000X},
	doi = {10.1109/MCS.2016.2558401},
}

@inproceedings{desaiControllingFormationsMultiple1998,
	title = {Controlling Formations of Multiple Mobile Robots},
	booktitle = {Proc. of the {{International Conference}} on {{Robotics}} and {{Automation}} (ICRA)},
	author = {Desai, J.P. and Ostrowski, J. and Kumar, V.},
	year = {1998},
	volume = {4},
	pages = {2864--2869},
	publisher = {IEEE},
	address = {Leuven, Belgium},
	doi = {10.1109/ROBOT.1998.680621},
	isbn = {978-0-7803-4300-9},
}

@article{dellaertFactorGraphsExploiting2021,
	title = {Factor {{Graphs}}: {{Exploiting Structure}} in {{Robotics}}},
	shorttitle = {Factor {{Graphs}}},
	author = {Dellaert, Frank},
	year = {2021},
	month = may,
	journal = {Annual Review of Control, Robotics, and Autonomous Systems},
	volume = {4},
	number = {1},
	pages = {141--166},
	issn = {2573-5144, 2573-5144},
	doi = {10.1146/annurev-control-061520-010504},
}

@article{roumeliotisDistributedMultirobotLocalization2002,
	title = {Distributed Multirobot Localization},
	author = {Roumeliotis, Stergios I. and Bekey, George A.},
	year = {2002},
	journal = {IEEE Transactions on Robotics and Automation},
	volume = {18},
	number = {5},
	pages = {781--795},
	publisher = {IEEE},
	issn = {1042296X},
	doi = {10.1109/TRA.2002.803461},
}

@inproceedings{leungDistributedDecentralizedCooperative2011,
	title = {Distributed and Decentralized Cooperative Simultaneous Localization and Mapping for Dynamic and Sparse Robot Networks},
	author = {Leung, Keith Y.K. and Barfoot, Timothy D. and Liu, Hugh H.T.},
	year = {2011},
	booktitle = {Proc of the {{International Conference on Robotics and Automation}} ({ICRA})},
	pages = {3841--3847},
	publisher = {IEEE},
	issn = {10504729},
	doi = {10.1109/ICRA.2011.5979783},
	isbn = {9781612843865},
}

@inproceedings{carrillo-arceDecentralizedMultirobotCooperative2013,
	title = {Decentralized Multi-Robot Cooperative Localization Using Covariance Intersection},
	author = {{Carrillo-Arce}, Luis C. and Nerurkar, Esha D. and Gordillo, Jose L. and Roumeliotis, Stergios I.},
	year = {2013},
	booktitle = {Proc. of the {{International Conference on Intelligent Robots and Systems}} ({{IROS}} } ,
	pages = {1412--1417},
	issn = {21530858},
	doi = {10.1109/IROS.2013.6696534},
	isbn = {9781467363587},
}

@article{cossetteDecentralizedStateEstimation2024,
	title = {Decentralized State Estimation: {{An}} Approach Using Pseudomeasurements and Preintegration},
	shorttitle = {Decentralized State Estimation},
	author = {Cossette, Charles Champagne and Shalaby, Mohammed Ayman and Saussi{\'e}, David and Forbes, James Richard},
	year = {2024},
	month = sep,
	journal = {The International Journal of Robotics Research},
	volume = {43},
	number = {10},
	pages = {1573--1593},
	issn = {0278-3649, 1741-3176},
	doi = {10.1177/02783649241230993},
}

@incollection{howardLocalizationMobileRobot2003,
	title = {Localization for {{Mobile Robot Teams}}: {{A Distributed MLE Approach}}},
	shorttitle = {Localization for {{Mobile Robot Teams}}},
	booktitle = {Experimental {{Robotics VIII}}},
	author = {Howard, Andrew and Matari{\'c}, Maja J. and Sukhatme, Gaurav S.},
	editor = {Siciliano, Bruno and Dario, Paolo},
	year = {2003},
	volume = {5},
	pages = {146--155},
	publisher = {Springer Berlin Heidelberg},
	address = {Berlin, Heidelberg},
	doi = {10.1007/3-540-36268-1_12},
	isbn = {978-3-540-00305-2},
}

@inproceedings{nerurkarDistributedMaximumPosteriori2009,
	title = {{Distributed Maximum a Posteriori Estimation for Multi-Robot Cooperative Localization}},
	author = {Nerurkar, Esha D. and Roumeliotis, Stergios I. and Martinelli, Agostino},
	year = {2009},
	booktitle = {Proc. of the {{International Conference on Robotics and Automation}} {(ICRA}},
	pages = {1402--1409},
	publisher = {IEEE},
	issn = {10504729},
	doi = {10.1109/ROBOT.2009.5152398},
}

@inproceedings{trawnyCooperativeMultirobotLocalization2009,
	title = {{Cooperative Multi-Robot Localization under Communication Constraints}},
	author = {Trawny, Nikolas and Roumeliotis, Stergios I. and Giannakis, Georgios B.},
	year = {2009},
	booktitle = {Proc. of the {{International Conference on Robotics and Automation}} {(ICRA}},
	pages = {4394--4400},
	publisher = {IEEE},
	issn = {10504729},
	doi = {10.1109/ROBOT.2009.5152606},
	isbn = {9781424427895},
}

@inproceedings{krickStabilizationInfinitesimallyRigid2008,
	title = {Stabilization of Infinitesimally Rigid Formations of Multi-Robot Networks},
	booktitle = {Proc. of the 47th {{IEEE Conference}} on {{Decision}} and {{Control}} ({CDC})},
	author = {Krick, Laura and Broucke, Mireille E. and Francis, Bruce A.},
	year = {2008},
	pages = {477--482},
	publisher = {IEEE},
	address = {Cancun, Mexico},
	doi = {10.1109/CDC.2008.4738760},
	isbn = {978-1-4244-3123-6},
}

@article{yuRigidGraphControl2008,
	title = {Rigid Graph Control Architecture for Autonomous Formations : {{Applying}} Classical Graph Theory to the Control of Multiagent Systems},
	author = {Yu, Changbin and Fidan, B. and Hendrickx, Julien},
	year = {2008},
	journal = {IEEE Control Systems Magazine},
	volume = {28},
	number = {6},
	pages = {46--63},
	issn = {1066-033X},
}

@phdthesis{babazadehDistanceBasedFormationControl2021,
	title = {Distance-{{Based Formation Control}} of {{Multi-Agent}}  {{Systems}}},
	author = {Babazadeh, Reza},
	year = {2021},
	month = dec,
	address = {Montreal, Quebec, Canada},
	school = {Concordia University},
}

@article{mukadamContinuoustimeGaussianProcess2018,
	title = {Continuous-Time {{Gaussian}} Process Motion Planning via Probabilistic Inference},
	author = {Mukadam, Mustafa and Dong, Jing and Yan, Xinyan and Dellaert, Frank and Boots, Byron},
	year = {2018},
	journal = {International Journal of Robotics Research},
	volume = {37},
	number = {11},
	eprint = {1707.07383},
	pages = {1319--1340},
	issn = {17413176},
	doi = {10.1177/0278364918790369},
	isbn = {0278364918},
}

@inproceedings{taFactorGraphApproach2014,
	title = {A Factor Graph Approach to Estimation and Model Predictive Control on {{Unmanned Aerial Vehicles}}},
	booktitle = {Proc. of the {{International Conference}} on {{Unmanned Aircraft Systems}} ({{ICUAS}})},
	author = {Ta, Duy-Nguyen and Kobilarov, Marin and Dellaert, Frank},
	year = {2014},
	month = may,
	pages = {181--188},
	publisher = {IEEE},
	address = {Orlando, FL, USA},
	doi = {10.1109/ICUAS.2014.6842254},
	isbn = {978-1-4799-2376-2},
}

@article{luskDistributedPipelineScalable2020,
	title = {A {{Distributed Pipeline}} for {{Scalable}}, {{Deconflicted Formation Flying}}},
	author = {Lusk, Parker C. and Cai, Xiaoyi and Wadhwania, Samir and Paris, Aleix and Fathian, Kaveh and How, Jonathan P.},
	year = {2020},
	month = oct,
	journal = {IEEE Robotics and Automation Letters},
	volume = {5},
	number = {4},
	pages = {5213--5220},
	issn = {2377-3766, 2377-3774},
	doi = {10.1109/LRA.2020.3006823},
}

@incollection{thakurSwarmInexpensiveHeterogeneous2021,
	title = {Swarm of {{Inexpensive Heterogeneous Micro Aerial Vehicles}}},
	booktitle = {Experimental {{Robotics}}},
	author = {Thakur, Dinesh and Tao, Yuezhan and Li, Rebecca and Zhou, Alex and Kushleyev, Alex and Kumar, Vijay},
	editor = {Siciliano, Bruno and Laschi, Cecilia and Khatib, Oussama},
	year = {2021},
	volume = {19},
	pages = {413--423},
	publisher = {Springer International Publishing},
	address = {Cham},
	doi = {10.1007/978-3-030-71151-1_37},
	isbn = {978-3-030-71151-1},
	langid = {english},
}

@article{xuOmniSwarmDecentralizedOmnidirectional2022,
	title = {Omni-{{Swarm}}: {{A Decentralized Omnidirectional Visual-Inertial-UWB State Estimation System}} for {{Aerial Swarms}}},
	author = {Xu, Hao and Zhang, Yichen and Zhou, Boyu and Wang, Luqi and Yao, Xinjie and Meng, Guotao and Shen, Shaojie},
	year = {2022},
	journal = {IEEE Transactions on Robotics},
	volume = {38},
	number = {6},
	eprint = {2103.04131},
	pages = {3374--3394},
	issn = {19410468},
	doi = {10.1109/TRO.2022.3182503},
}

@article{nguyenPersistentlyExcitedAdaptive2020,
	title = {Persistently {{Excited Adaptive Relative Localization}} and {{Time-Varying Formation}} of {{Robot Swarms}}},
	author = {Nguyen, Thien Minh and Qiu, Zhirong and Nguyen, Thien Hoang and Cao, Muqing and Xie, Lihua},
	year = {2020},
	journal = {IEEE Transactions on Robotics},
	volume = {36},
	number = {2},
	pages = {553--560},
	publisher = {IEEE},
	issn = {19410468},
	doi = {10.1109/TRO.2019.2954677},
}

@article{caoSimilarFormationControl2023,
	title = {Similar {{Formation Control}} via {{Range}} and {{Odometry Measurements}}},
	author = {Cao, Kun and Cao, Muqing and Xie, Lihua},
	year = {2023},
	journal = {IEEE Transactions on Cybernetics},
	pages = {1--12},
	publisher = {IEEE},
	issn = {21682275},
	doi = {10.1109/TCYB.2023.3263475},
}

@article{vanderhelmOnboardRangebasedRelative2020a,
	title = {On-Board Range-Based Relative Localization for Micro Air Vehicles in Indoor Leader--Follower Flight},
	author = {Van Der Helm, Steven and Coppola, Mario and McGuire, Kimberly N. and De Croon, Guido C. H. E.},
	year = {2020},
	month = mar,
	journal = {Autonomous Robots},
	volume = {44},
	number = {3-4},
	pages = {415--441},
	issn = {0929-5593, 1573-7527},
	doi = {10.1007/s10514-019-09843-6},
	langid = {english},
}

@inproceedings{zhangEfficientRangeConstraintManifold2022,
	title = {Efficient {{Range-Constraint Manifold Optimization}} with {{Application}} to {{Cooperative Navigation}}},
	author = {Zhang, Yetong and Chen, Gerry and Rutkowski, Adam and Dellaert, Frank},
	year = {2022},
	booktitle = {Proc. of the {{International Conference on Intelligent Robots and Systems}}({IROS})},
	volume = {2022-Octob},
	pages = {9950--9956},
	publisher = {IEEE},
	issn = {21530866},
	doi = {10.1109/IROS47612.2022.9982188},
}

@inproceedings{cunninghamDDFSAMConsistentDistributed2013a,
	title = {{{DDF-SAM}} 2.0: {{Consistent}} Distributed Smoothing and Mapping},
	author = {Cunningham, Alexander and Indelman, Vadim and Dellaert, Frank},
	year = {2013},
	booktitle = {Proc. of the {{International Conference on Robotics and Automation}({ICRA})}},
	pages = {5220--5227},
	publisher = {IEEE},
	issn = {10504729},
	doi = {10.1109/ICRA.2013.6631323},
}

@software{gtsam,
	author       = {Frank Dellaert and GTSAM Contributors},
	title        = {borglab/gtsam},
	month        = May,
	year         = 2022,
	publisher    = {Georgia Tech Borg Lab},
	version      = {4.2a8},
	doi          = {10.5281/zenodo.5794541},
	howpublished = {\url{https://github.com/borglab/gtsam}}
}

@inproceedings{turpinDecentralizedFormationControl2012,
	title = {Decentralized Formation Control with Variable Shapes for Aerial Robots},
	booktitle = {Proc. of the {{International Conference}} on {{Robotics}} and {{Automation}}},
	author = {Turpin, Matthew and Michael, Nathan and Kumar, Vijay},
	year = {2012},
	month = may,
	pages = {23--30},
	publisher = {IEEE},
	address = {St Paul, MN, USA},
	doi = {10.1109/ICRA.2012.6225196},
}

@article{zhouSwarmMicroFlying2022,
	author = {Xin Zhou  and Xiangyong Wen  and Zhepei Wang  and Yuman Gao  and Haojia Li  and Qianhao Wang  and Tiankai Yang  and Haojian Lu  and Yanjun Cao  and Chao Xu  and Fei Gao },
	title = {Swarm of micro flying robots in the wild},
	journal = {Science Robotics},
	volume = {7},
	number = {66},
	year = {2022},
	doi = {10.1126/scirobotics.abm5954}
}

@article{nguyenDistanceBasedCooperativeRelative2019,
	title = {Distance-{{Based Cooperative Relative Localization}} for {{Leader-Following Control}} of {{MAVs}}},
	author = {Nguyen, Thien-Minh and Qiu, Zhirong and Nguyen, Thien Hoang and Cao, Muqing and Xie, Lihua},
	year = {2019},
	month = oct,
	journal = {IEEE Robotics and Automation Letters},
	volume = {4},
	number = {4},
	pages = {3641--3648},
	issn = {2377-3766, 2377-3774},
	doi = {10.1109/LRA.2019.2926671},
}

@article{goudarRangeVisualInertialSensorFusion2024,
	title = {Range-{{Visual-Inertial Sensor Fusion}} for {{Micro Aerial Vehicle Localization}} and {{Navigation}}},
	author = {Goudar, Abhishek and Zhao, Wenda and Schoellig, Angela P.},
	year = {2024},
	month = jan,
	journal = {IEEE Robotics and Automation Letters},
	volume = {9},
	number = {1},
	pages = {683--690},
	issn = {2377-3766, 2377-3774},
	doi = {10.1109/LRA.2023.3335772},
}

\end{document}